\documentclass[runningheads]{llncs}
\usepackage[T1]{fontenc}
\usepackage{graphicx}
\usepackage{comment}
\usepackage{amsmath}
\usepackage{tabularray}
\usepackage{multicol}
\usepackage{multirow}
\usepackage{amssymb}
\usepackage{bm}

\begin{document}

\title{Towards a text-based quantitative and explainable histopathology image analysis}
\titlerunning{Text-based quantitative and explainable histopathology image analysis}
\author{Anh Tien Nguyen \and Trinh Thi Le Vuong \and  Jin Tae Kwak}
\authorrunning{A.T. Nguyen \& T.T.L. Vuong \&  J.T. Kwak}
\institute{School of Electrical Engineering, Korea University, Seoul 02841, Korea \\
\email{\{ngtienanh,trinhvg,jkwak\}@korea.ac.kr}}

\maketitle
\begin{abstract}
    Recently, vision-language pre-trained models have emerged in computational pathology. Previous works generally focused on the alignment of image-text pairs via the contrastive pre-training paradigm. Such pre-trained models have been applied to pathology image classification in zero-shot learning or transfer learning fashion. Herein, we hypothesize that the pre-trained vision-language models can be utilized for quantitative histopathology image analysis through a simple image-to-text retrieval. To this end, we propose a Text-based Quantitative and Explainable histopathology image analysis, which we call TQx. Given a set of histopathology images, we adopt a pre-trained vision-language model to retrieve a word-of-interest pool. The retrieved words are then used to quantify the histopathology images and generate understandable feature embeddings due to the direct mapping to the text description. To evaluate the proposed method, the text-based embeddings of four histopathology image datasets are utilized to perform clustering and classification tasks. The results demonstrate that TQx is able to quantify and analyze histopathology images that are comparable to the prevalent visual models in computational pathology. The repository is available at \url{https://github.com/QuIIL/TQx}.

\end{abstract}

\section{Introduction}
    Over the past years, the success of deep learning models has been attributable to the single-modal pre-trained models in computer vision and natural language processing. In computer vision, various convolutional neural networks (CNNs) \cite{resnet,efficientnetv2,convnext} and Vision Transformer (ViT) \cite{vit,swin2,maxvit}, known as vision models, are pre-trained on ImageNet and have been successfully applied to a wide range of downstream tasks such as image classification, retrieval, and segmentation~\cite{Deng_2023_ICCV}. In natural language processing, several pre-trained language models such as BERT~\cite{bert} and GPT~\cite{gpt} are available and have shown to be effective in many text-based applications, including text classification, question answering, translation, and summarization~\cite{min2023recent}.
    Recently, there exists an emerging practice of fusing the vision and natural language models, which forms vision-language models (VLMs). VLMs are, in general, pre-trained on an extensive collection of image-text datasets by jointly learning visual and textual modalities/knowledge. A notable example is CLIP~\cite{clip}, which simultaneously optimizes vision and language models to align the visual and textual representations. Computational pathology is not an exception. PLIP~\cite{plip} and QUILT-Net~\cite{quilt} are two exemplary VLMs that were pre-trained on massive histopathology datasets with pairs of histopathology image and histopathology description via the contrastive learning paradigm similar to CLIP. 
    Such VLMs build a solid foundation for various types of downstream tasks. In particular, these models have been successfully applied to several image classification tasks such as lymph-node metastasis detection, tissue phenotyping, and Gleason grading \cite{plip,quilt} without further training or fine-tuning, i.e., zero-shot image classification.
    
    We have observed that the existing works on VLMs have mainly focused on the pre-training process and straightforward, direct application to downstream tasks \cite{clip,plip,quilt,align}. Some have sought to utilize VLMs with transfer learning and knowledge distillation \cite{biomedclip,dai2022enabling}. These approaches are, by and large, similar to the way the pre-trained vision models and language models are used, which do not fully explore the potential of VLMs. Herein, we hypothesize that the pre-trained VLMs \textit{per se} are capable of conducting quantitative histopathology image analysis. In other words, VLMs are able to quantify histopathology images, and the resultant quantitative features can be used for downstream tasks. To test our hypothesis, we propose a Text-based Quantitative and Explainable histopathology image analysis framework, called TQx. 
    We systematically evaluate the effectiveness of TQx using four histopathology image datasets by performing clustering and classification tasks. The experimental results suggest that TQx not only provides a capacity for quantifying and analyzing histopathology images comparable to the conventional vision models but also permits the direct interpretation of the results with human-readable words.

\section{Methodology}
\begin{figure*}[!]
            \includegraphics[width=\textwidth]{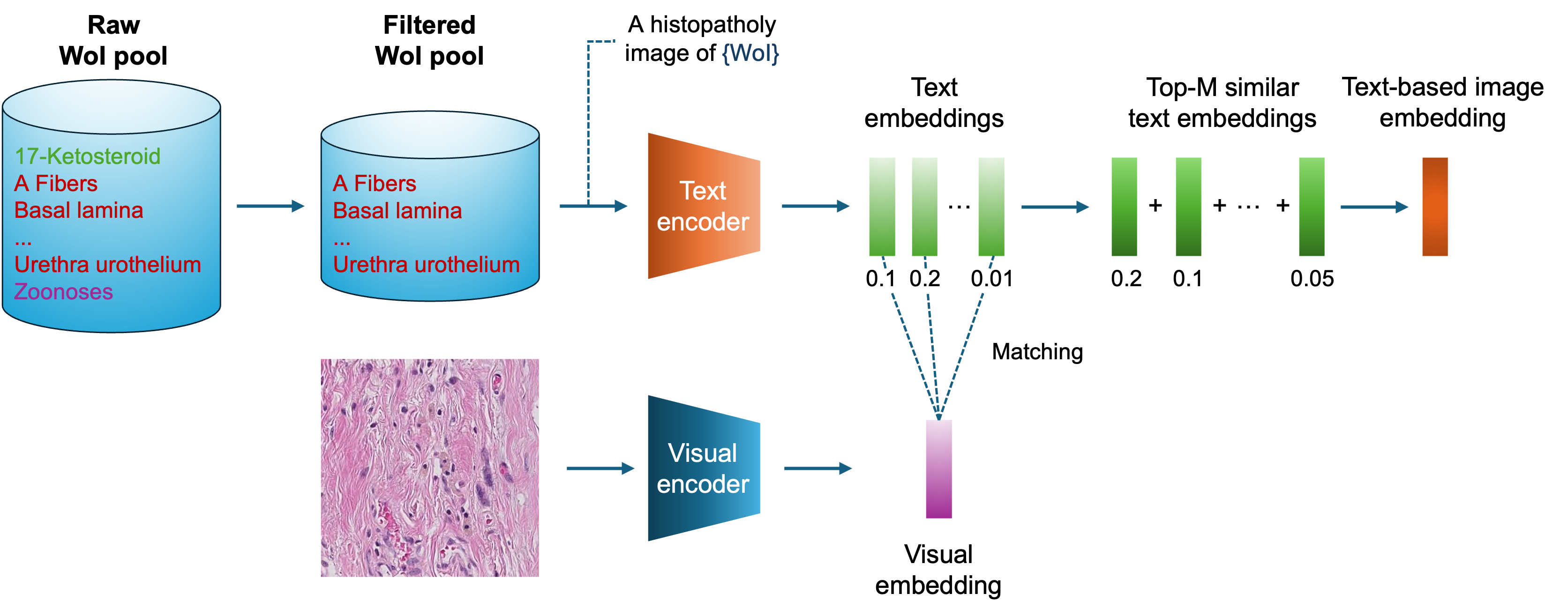}
            \caption{The raw WoI pool stores all UMLS~\cite{umls} pathology terms of various semantic types. The filtered pool is obtained by selecting a particular semantic type under consideration. The pair of encoders from a pre-trained VLM generates text and visual embeddings, which are then compared together. The similarity scores from the comparison are normalized and then used as weights to produce a text-based embedding.} \label{TQx}
        \end{figure*}

    TQx involves two major components: 1) a pre-trained VLM and 2) a word-of-interest (WoI) pool. The VLM has a text encoder and a visual encoder that was jointly optimized via contrastive learning~\cite{clip}. The WoI pool includes a set of pathology terms that explain the characteristics of histopathology images. Overall, by utilizing cosine similarity, the VLM retrieves relevant keywords from the WoI pool, then the text-based image embedding is generated from these keywords (Fig. \ref{TQx}). The most important step is to filter the related terms because they directly affect the generation of the text-based representation. Therefore, in the following sections, we will explain this procedure in detail.

    \subsection{Text-based image representation}
        Suppose that we are given a pre-trained VLM $\mathcal{V}$, a set of pathology images $\mathcal{X} = \{x_i\}_{i=1}^{N}$, and a WoI pool $\mathcal{W} = \{w_j\}_{j=1}^{N_w}$ where $x_i$ is the $i$th histopathology image, $w_j$ is the $j$th keyword, and $N$ and $N_w$ are the number of histopathology images and keywords, respectively. 
        Each $x_i$ undergoes the visual encoder of $\mathcal{V}$ to produce the visual embedding $\mathbf{v}_i$, and each $w_j$ is fed into the text encoder of $\mathcal{V}$ to generate the corresponding text embedding $\mathbf{f}_j$. 
        For all $(\mathbf{v}_i,\mathbf{f}_j)$, we compute the similarity scores, producing $\mathcal{S} = \{s_{i,j} | i=1,...,N \wedge j=1,...,N_w\}$ where $s_{i,j}$ denotes the similarity between $x_i$ and $w_j$. 
        Using $\mathcal{S}$, we select the top-M keywords $\mathcal{W}^{M} = \{w'_i\}_{i=1}^{M}$ that are most representative of $\mathcal{X}$. For each image $x_i$, we first compute the rank of the keywords in $\mathcal{W}$ as follows: $\forall (j,k)$, $r_{i,j} < r_{i,k}$ if $s_{i,j} < s_{i,k}$ where $r_{i,j} \in \{1,...,N_w\}$ is the rank of $w_j$ for $x_i$. Then, we average the rank of the keywords among $\mathcal{X}$ to select a set of $M$ keywords with the highest ranks, designated as $\mathcal{W}^{M} = \{w'_i\}_{i=1}^{M}$, and use it to produce the corresponding text embeddings $\mathcal{F}^{M} = \{\mathbf{f}'_j\}_{j=1}^{M}$ and their similarity scores $\mathcal{S}^{M} = \{s'_{i,j} | i=1,...,N \wedge j=1,...,M\}$.
        
        To obtain the text-based image representation for each $x_i$, we normalize the similarity scores between $x_i$ and all the keywords in $\mathcal{W}^{M}$, i.e., $\{s'_{i,j}\}_{j=1}^{M}$, using the softmax operation to produce weights for the text embeddings $\{\alpha_{j}\}_{j=1}^{M}$. Then, we compute a weighted sum of the text embeddings to generate the final text-based image embedding of $x_i$ given by $\mathbf{f}^{T}_{i} = \sum_{j=1}^{M} \alpha_j \mathbf{f}'_j$. 
        The text-based image embedding $\mathbf{f}^{T}$ delivers the abstract textual information of the input image, which can be used for quantitative analysis. The top-M keywords $\mathcal{W}^{M}$ are human-readable and -understandable, thus enabling the interpretation of the results. Therefore, the text-based image embedding is self-explainable \textit{per se}.
        
    \subsection{Construction of word-of-interest pool}
        To construct the WoI pool, we utilize QUILT-1M~\cite{quilt}, the largest image-text dataset in pathology containing a wide range of pathology keywords. First, we collect all keywords (called entities) from QUILT-1M, then cross-check and process with standard terms in the Unified Medical Language System (UMLS)~\cite{umls}. After eliminating the duplicate terms using the UMLS concept unique identifier (CUI), we obtain the raw WoI pool of 28,292 keywords, which is designated as $\mathcal{W}_{Level-0}$.

    \subsection{Refinement of word-of-interest pool}
        In the raw WoI pool $\mathcal{W}_{Level-0}$, each keyword is associated with the corresponding UMLS semantic type that defines its category, such as \textit{Tissue}, \textit{Cell}, or \textit{Neoplastic Process}. In the UMLS, these entities are categorized based on the tree-like semantic network, and thus, one semantic type may include one or more sub-types or -groups. For example, \textit{Fully Formed Anatomical Structure} contains multiple smaller sub-groups such as \textit{Cell}, \textit{Tissue}, and \textit{Cell Component}. Each semantic (sub-)type has its own histopathology meaning.
        
        To investigate the impact of the WoI pool and the semantic category on TQx, we choose three UMLS semantic types that are relevant to pathology diagnosis: 1) \textit{Neoplastic Process}: abnormal growth of tissue, 2) \textit{Disease or Syndrome}: an abnormal condition of an organism, and 3) \textit{Pathologic Function}: a disordered process, activity, or state of the organism or part of it.
        Then, we construct three WoI pools including $\mathcal{W}_{Level-3}$ (\textit{Neoplastic Process}), $\mathcal{W}_{Level-2}$ (\textit{Disease or Syndrome}), and $\mathcal{W}_{Level-1}$ (\textit{Pathologic Function}) with 2,215, 5,441, and 6,232 keywords, respectively. In total, we build four WoI pools. For each of the four WoI pools, we retrieve the top-M highly ranked keywords with $M=1000$, resulting in $\mathcal{W}_{Level-0}^{M}$, $\mathcal{W}_{Level-1}^{M}$, $\mathcal{W}_{Level-2}^{M}$, and $\mathcal{W}_{Level-3}^{M}$. $\mathcal{W}_{Level-0}^{M}$ (raw WoI pool) is the most inclusive and general while $\mathcal{W}_{Level-3}^{M}$ (\textit{Neoplastic Process}) is the most confined and specific. Moreover, using the four WoI pools, we generate the four sets of text-based feature embeddings: $\mathcal{F}^{\mathcal{W}_{Level-0}^{M}}$, $\mathcal{F}^{\mathcal{W}_{Level-1}^{M}}$, $\mathcal{F}^{\mathcal{W}_{Level-2}^{M}}$, and $\mathcal{F}^{\mathcal{W}_{Level-3}^{M}}$.
        
    \subsection{Pre-trained vision-language model}
        We adopt QUILT-Net~\cite{quilt} as the VLM $\mathcal{V}$ for histopathology image analysis due to its superior performance compared to state-of-the-art VLMs such as CLIP and PLIP. In QUILT-Net, the visual encoder is constructed based on ViT-B/32~\cite{vit}, which splits input images into 32x32 tiles and forwards through 12 self-attention layers with 12 heads. The text encoder inherits an architecture of GPT-2~\cite{gpt2} with 12 self-attention layers with 8 heads. These encoders are initialized with the weights of CLIP and then fine-tuned with QUILT-1M.
        
\section{Experiments}
    \subsection{Datasets}
        We employ four different public datasets to analyze the effectiveness of TQx. 
        The first dataset, called \textbf{Colon}~\cite{joint}, is a colorectal cancer grading dataset, which includes 9,857 patches of size 512 x 512 with four classes: \textit{benign} (BN), \textit{well-differentiated cancer} (WD), \textit{moderately differentiated cancer} (MD), and \textit{poorly differentiated cancer} (PD).
        \textbf{WSSS4LUAD}~\cite{wsss} is the second dataset for lung cancer detection. We extracted 3,526 patches of size 224 × 224 that are labeled as \textit{normal} (NOR) or \textit{tumor} (TUM).
        The third dataset is \textbf{BACH}~\cite{bach} that contains 58,539 images of size 1024 x 1024 for breast cancer staging with four categories: \textit{normal} (NOR), \textit{benign} (BN), \textit{in situ carcinoma} (SITU), and \textit{invasive carcinoma} (IVS). 
        The last dataset \textbf{Bladder}~\cite{bladder}  is designed for bladder cancer grading, comprising 14,258 patches of size 512 × 512 that are annotated as \textit{normal} (NOR), \textit{low-grade cancer} (LOW), and \textit{high-grade cancer} (HIGH).
        
        \begin{figure*}[t!]
            \includegraphics[width=\textwidth]{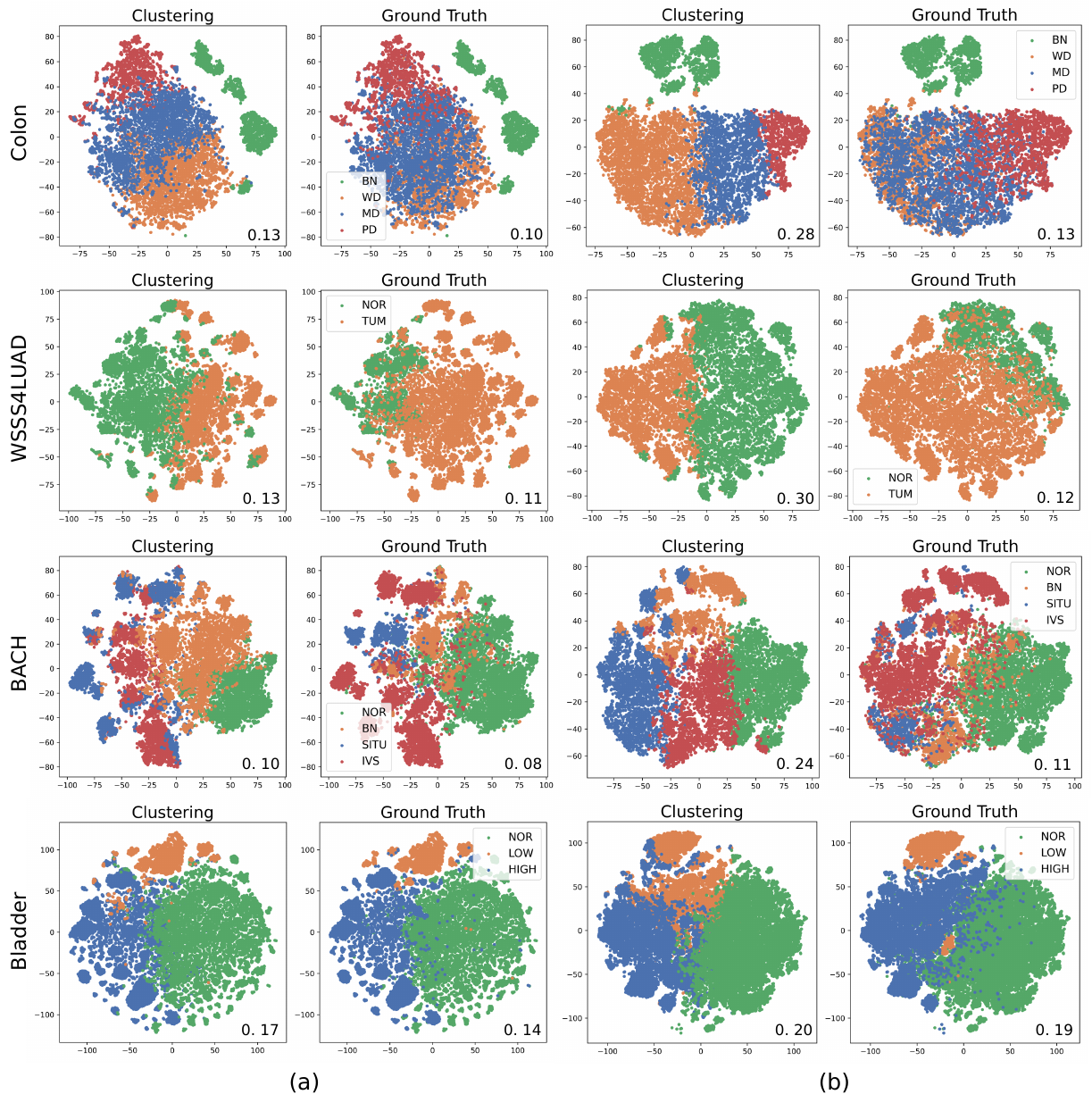}
            \caption{Clustering results with $\mathcal{W}_{Level-3}^{M}$ (\textit{Neoplastic Process}) of (a) visual embeddings and (b) text-based image embeddings. In the \textit{Ground Truth} plots, samples are re-assigned to the clusters using the ground truth class labels. The bottom numbers show silhouette coefficients measuring how similar an embedding is to its own cluster.} \label{clustering}
        \end{figure*}

    \subsection{Analysis of text-based image representation}
        To investigate the effectiveness of the text-based image representation, we conduct two tasks: 1) Clustering and 2) Classification. Both tasks are evaluated for four aforementioned WoI pools. 
        To compare between visual and text-based embeddings, we adopt a pair of visual and textual encoders from QUILT-Net~\cite{quilt}. For \textbf{clustering}, we employ Lloyd's K-Means method~\cite{kmeans}, where K is set to a number of classification classes. K-Means++ is used for cluster initialization, with the maximum number of iterations is set to 300. We evaluate the clustering quality using silhouette coefficients and further assess the results visually through t-SNE~\cite{tsne}.

        For \textbf{classification}, we construct a simple multi-layered perceptron with two fully-connected layers, a ReLU activation function, and a batch normalization layer. The classifier is trained for 300 epochs using Adam optimizer with a learning rate 0.01. 
        For Colon, Bladder, and BACH, four evaluation metrics are employed: accuracy ($Acc$), accuracy cancer ($Acc_c$), macro-averaged F1 score ($F1$), and quadratic-weighted kappa score ($K_w$). 
        For WSSS4LUAD, $Acc$, $F1$, precision ($Pre$), and recall ($Rec$) are used. 
        The experiments are replicated with 50 different initialization seeds to calculate the mean and standard deviation of each evaluation metric.

        \begin{table}[t!]
        \centering
        \caption{Silhouette coefficients of clustering with visual embedding and text-based embedding from four different WoI pools.}\label{silhouette}
            \resizebox{0.8\textwidth}{!}{
            \begin{tabular}{llclclclc}
                \hline
                Embedding & \multicolumn{1}{c}{} & Colon~\cite{joint} & \multicolumn{1}{c}{} & WSSS4LUAD~\cite{wsss} & \multicolumn{1}{c}{} & BACH~\cite{bach} & \multicolumn{1}{c}{} & Bladder~\cite{bladder} \\ \hline
                Visual                         &                      & 0.13  &                      & 0.13      &                      & 0.10 &                      & 0.17    \\
                Text - $\mathcal{W}_{Level-0}^{M}$                     &                      & 0.25  &                      & 0.26      &                      & 0.21 &                      & 0.27    \\
                Text - $\mathcal{W}_{Level-1}^{M}$                     &                      & 0.27  &                      & 0.30      &                      & 0.25 &                      & 0.25    \\
                Text - $\mathcal{W}_{Level-2}^{M}$                     &                      & 0.28  &                      & 0.29      &                      & 0.25 &                      & 0.23    \\
                Text - $\mathcal{W}_{Level-3}^{M}$                     &                      & 0.28  &                      & 0.30      &                      & 0.24 &                      & 0.20    \\ \hline
                \end{tabular}}
        \end{table}
        
\section{Results and Discussion}
    \subsection{Clustering}
        Fig. \ref{clustering} visualizes the clustering results for $\mathcal{W}_{Level-3}^{M}$ using t-SNE, and Table \ref{silhouette} shows the silhouette coefficients of the four datasets. The results show that the text-based image embeddings well formed the clusters corresponding to the ground truth class labels across the datasets. It was striking that the silhouette coefficients of the text-based image embeddings were consistently larger than those of the visual embeddings such that $\ge$0.12, $\ge$0.13, $\ge$0.11, and $\ge$0.03 for Colon, WSSS4LUAD, BACH, and Bladder, respectively. Among the text-based image embeddings, $\mathcal{F}^{\mathcal{W}_{Level-0}^{M}}$ generally obtained the worst results, whereas others were comparable to each other.

        \begin{figure*}[t!]
            \includegraphics[width=\textwidth]{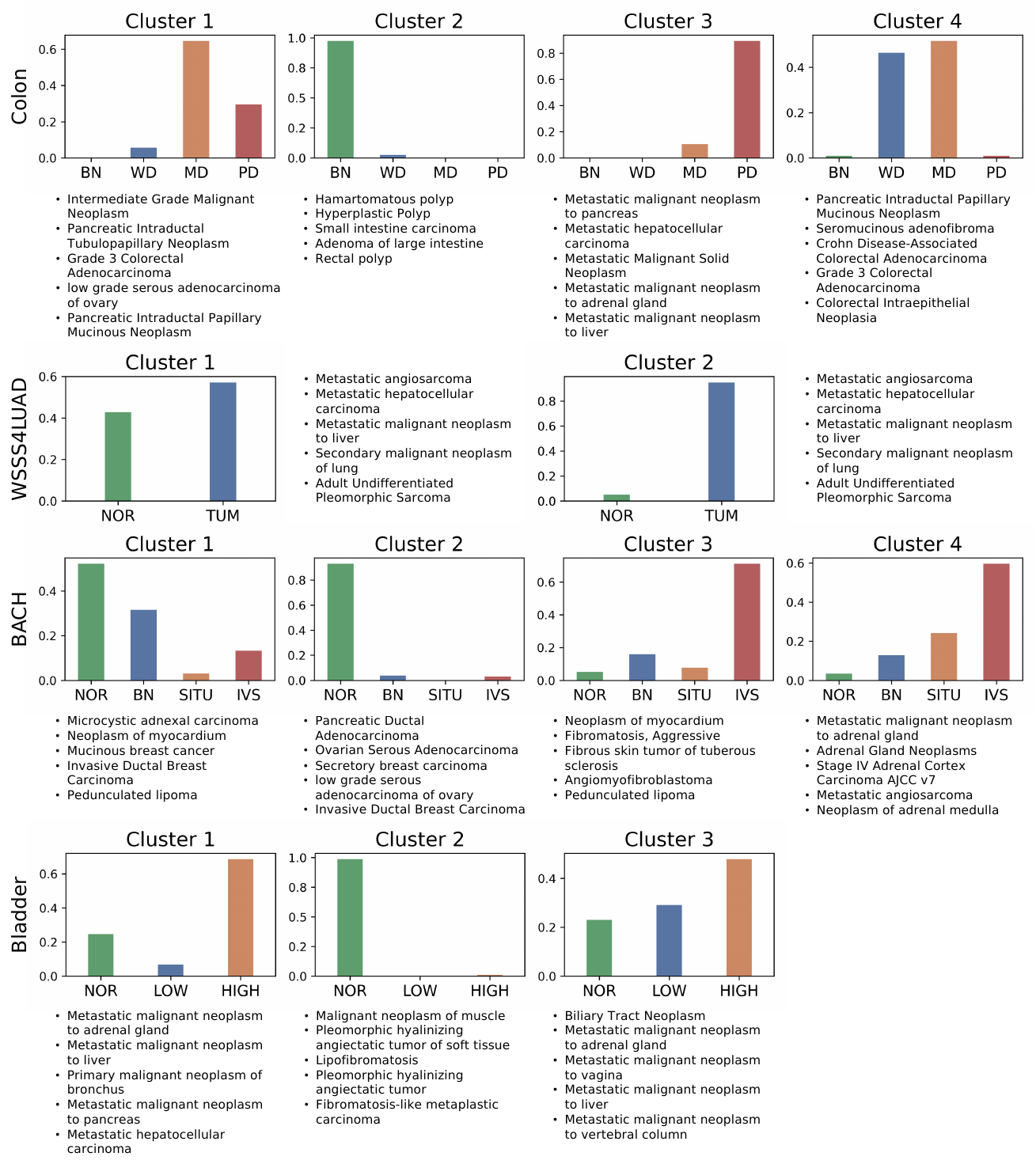}
            \caption{The bar plots show the percentage of samples per class in each cluster, based on the clustering with $\mathcal{W}_{Level-3}^{M}$ (\textit{Neoplastic Process}). Five keywords with the highest average ranks are shown next to the corresponding bar plot.} \label{common_terms}
        \end{figure*}
        
        Fig. \ref{common_terms} depicts the percentage of data samples per class and the top-5 keywords within each cluster. We made the following observations. Each cluster was dominated by one or two categories that are related to each other; for example, Colon Cluster 4 was dominated by WD and MD, and TUM was predominant in WSSS4LUAD Cluster 2. Non-cancerous samples were clearly separated from cancerous samples across the four datasets; for example, most of BN belonged to Cluster 2 for Colon, and most of NOR and BN resided in Cluster 1 and Cluster 2 for BACH, respectively. 
        Furthermore, we analyzed the top-5 matching keywords per cluster. The keywords generally matched the type of dataset. For instance, in Colon, Cluster 2, dominated by BN, was associated with \textit{Hamartomatous polyp} and \textit{Hyperplastic polyp} that are non-cancerous, and other clusters (Cluster 1, 3, and 4), mainly containing cancers, were matched with cancer-related keywords such as \textit{Grade 3 Colorectal Adenocarcinoma} and \textit{Colorectal Intraepithelial Neoplasia}. 
        For WSSS4LUAD, BACH, and Bladder, most of the clusters were shown to be highly relevant to cancer-related terms. Some of them are specific to the original organ, such as \textit{Secondary malignant neoplasm of lung} for WSSS4LUAD and \textit{Invasive Ductal Breast Carcinoma} for BACH. Some others were related to different organs; for example, \textit{Pancreatic Ductal Adenocarcinoma} for BACH and \textit{Metastatic malignant neoplasm to adrenal gland} for Bladder. Though cancer-related terms were prevalent, the clusters representing BN and/or NOR were often aligned with benign, slow-growing tumors such as \textit{Pleomorphic hyalinizing angiectatic tumor} (slow-growing tumor) and \textit{Lipofibromatosis} (benign soft tissue tumor).

    \subsection{Classification}
        The classification results on the four datasets are shown in Table \ref{colon_luad_result}. 
        For Colon and BACH, the visual embeddings performed best, outperforming the text-based image embeddings over the four evaluation metrics in Colon and three evaluation metrics ($Acc$, $Acc_c$, and $F1$) in BACH. For WSSS4LUAD, $\mathcal{F}^{\mathcal{W}_{Level-3}^{M}}$ achieved the best performance in $Acc$, $Acc_c$, and $F1$. As for Bladder, the visual embeddings and text-based image embeddings were comparable to each other; the visual embeddings obtained the best $Acc$ and $K_w$, while the highest $Acc_c$ and $F1$ were attained by the text-based image embeddings ($\mathcal{F}^{\mathcal{W}_{Level-0}^{M}}$).

        Among the four types of the text-based image embeddings, $\mathcal{F}^{\mathcal{W}_{Level-3}^{M}}$, in general, achieved the best performance and $\mathcal{F}^{\mathcal{W}_{Level-0}^{M}}$ was inferior to others except for Bladder. Hence, the more specific the WoI pool is, the better performance we tend to obtain. This indicates that the performance of the embeddings depends on the selection of the WoI pool, and the well-defined WoI pool could further improve the histopathology image analysis by instructing the specific patterns in histopathology images. 

        \begin{table}[t!]
            \centering
            \caption{Classification results using visual and text-based image embeddings.}\label{colon_luad_result}
            \resizebox{\textwidth}{!}{
            \begin{tabular} {lccccccccccccccccc}
                \hline
                \multirow{2}{*}{Embedding} & & \multicolumn{7}{c}{Colon~\cite{joint}}    &  & \multicolumn{7}{c}{WSSS4LUAD~\cite{wsss}} \\ 
                 & & $Acc$ (\%) & &  $Acc_c$ (\%) && $F1$     && $K_w$  &  & $Acc$ (\%) &&  $Pre$ && $F1$   && $Rec$ \\            
                 \cline{1-17}
                Vision & & $\bm{\textbf{81.7$\pm$1.8}}$ & &\textbf{74.6$\pm$2.5} & &\textbf{0.778$\pm$0.014}  &&  \textbf{0.913$\pm$0.008} &
                & 83.9$\pm$1.7 && 0.615$\pm$0.016 && 0.640$\pm$0.022 && 0.889$\pm$0.007   \\
    
                Text - $\mathcal{W}_{Level-0}^{M}$ & & 77.0$\pm$2.7 && 68.7$\pm$4.2 && 0.729$\pm$0.019  &&  0.895$\pm$0.011
                & & 87.3$\pm$0.4 && 0.639$\pm$0.003 && 0.681$\pm$0.005 & &\textbf{0.919$\pm$0.003} \\
    
                Text - $\mathcal{W}_{Level-1}^{M}$ & & 78.3$\pm$2.2 && 70.0$\pm$3.1 && 0.743$\pm$0.017  &&  0.901$\pm$0.009 &  
                & 87.4$\pm$0.8 && 0.639$\pm$0.006 && 0.681$\pm$0.009 && 0.911$\pm$0.007   \\ 
    
                Text - $\mathcal{W}_{Level-2}^{M}$ & & 78.5$\pm$1.7 && 70.2$\pm$2.3 && 0.748$\pm$0.012  &&  0.902$\pm$0.007 & 
                & 87.0$\pm$1.2 && 0.637$\pm$0.009 && 0.677$\pm$0.015 && 0.913$\pm$0.009   \\ 
                
                Text - $\mathcal{W}_{Level-3}^{M}$ & & 79.6$\pm$2.5 && 71.8$\pm$3.7 && 0.753$\pm$0.021  &&  0.904$\pm$0.013 &
                & \textbf{89.3$\pm$0.5} && \textbf{0.645$\pm$0.005} && \textbf{0.691$\pm$0.007} && 0.866$\pm$0.004   \\  

                \cline{1-17}
                \multirow{2}{*}{Embedding} & & \multicolumn{7}{c}{BACH~\cite{bach}}    &  & \multicolumn{7}{c}{Bladder~\cite{bladder}} \\ 
                 & & $Acc$ (\%) &&  $Acc_c$ (\%) && $F1$     && $K_w$  &  & $Acc$ (\%) &&  $Acc_c$ (\%) && $F1$   && $K_w$ \\            
                 \cline{1-17}
                Vision & & \textbf{75.1$\pm$1.7} && \textbf{64.8$\pm$2.4} && \textbf{0.660$\pm$0.020}  &&  0.762$\pm$0.044 &
                & \textbf{86.1$\pm$0.9} && 85.7$\pm$1.0 && 0.726$\pm$0.014 && \textbf{0.749$\pm$0.016}   \\
    
                Text - $\mathcal{W}_{Level-0}^{M}$ & & 72.7$\pm$0.6 && 58.1$\pm$0.8 && 0.603$\pm$0.016  &&  0.765$\pm$0.008
                & & 85.2$\pm$0.8 && \textbf{88.9$\pm$0.3} && \textbf{0.758$\pm$0.009} && 0.716$\pm$0.013   \\
    
                Text - $\mathcal{W}_{Level-1}^{M}$ & & 71.9$\pm$1.1 && 56.0$\pm$1.9 && 0.569$\pm$0.013  &&  0.765$\pm$0.017 &  
                & 82.0$\pm$0.8 && 83.8$\pm$0.3 && 0.609$\pm$0.013 && 0.679$\pm$0.016   \\
    
                Text - $\mathcal{W}_{Level-2}^{M}$ & & 72.3$\pm$1.4 && 57.0$\pm$2.1 && 0.574$\pm$0.011  &&  0773$\pm$0.013 & 
                & 81.2$\pm$0.8 && 84.0$\pm$0.3 && 0.604$\pm$0.015 && 0.662$\pm$0.013   \\
                
                Text - $\mathcal{W}_{Level-3}^{M}$ & & 73.1$\pm$2.1 && 58.6$\pm$1.3 && 0.564$\pm$0.017  &&  \textbf{0.801$\pm$0.013} &
                & 78.2$\pm$0.4 && 88.3$\pm$0.4 && 0.683$\pm$0.010 && 0.594$\pm$0.005   \\
                  \hline
            \end{tabular}}
        \end{table}

\section{Conclusions}
    Herein, we propose TQx, a text-based quantitative and explainable histopathology image analysis framework that exploits a pre-trained VLM and a simple image-to-text retrieval. The text-based image embeddings, driven by TQx, can be used for quantitative analysis and are directly associated with the histopathology terms, which are human-readable and -understandable without any post-processing or interpretation. The future study will further investigate the construction and optimization of the WoI pool and the application of the text-based image embeddings for other downstream tasks.

\begin{credits}
    \subsubsection{\ackname} This study was supported by the National Research Foundation of Korea (NRF) (No. 2021R1A2C2014557) and by the Ministry of Trade, Industry and Energy(MOTIE) and Korea Institute for Advancement of Technology (KIAT) through the International Cooperative R\&D program (P0022543).
    
    \subsubsection{\discintname}
    There are no competing interests.
\end{credits}

\bibliographystyle{splncs04}
\bibliography{references}

\end{document}